\documentclass[conference]{IEEEtran}
\IEEEoverridecommandlockouts
\usepackage{cite}
\usepackage{amsmath,amssymb,amsfonts}
\usepackage{algorithmic}
\usepackage{graphicx}
\usepackage{textcomp}
\usepackage{xcolor}
\usepackage{booktabs}
\usepackage{bm}
\usepackage{pifont}
\usepackage{fvextra}
\usepackage{multirow}

\begin{document}

\title{In-Context Adaptation of Encoder-Decoder Models in Speech Recognition
}

\author{

\IEEEauthorblockN{Yen Meng \qquad Sharon Goldwater \qquad Hao Tang}
\IEEEauthorblockA{\textit{The Centre for Speech Technology Research, University of Edinburgh, United Kingdom} \\
yen.meng@ed.ac.uk, sgwater@inf.ed.ac.uk, hao.tang@ed.ac.uk}

}

\maketitle

\begin{abstract}
In-context learning offers an appealing approach to adapt automatic speech
recognition (ASR) models to new speakers, accents, and domains by providing
speech-text pairs as demonstrations at inference time.
Recent work shows that some LLM-based speech models
are capable of ASR in-context adaptation,
when providing interleaved speech-text demonstrations.
In this work, we ask whether in-context adaptation is an
inherent ability for all encoder-decoder models.
We study two forms of demonstration, collated and interleaved demonstration,
across six encoder-decoder models,
spanning conventional cross-attention-based and LLM-based
architectures.
We find that all tested models are able to perform in-context adaptation
out of the box,
achieving up to 30\% relative improvement in the oracle experiments
and up to 23\% using first-pass hypotheses. 
Through controlled experiments on three English datasets, we show that lexical and speaker information both contribute to successful adaptation.
While interleaved demonstration is effective in certain cases,
collated demonstration brings consistent adaptation across the board.
Our results suggest that in-context adaptation for ASR is not unique to specific architectures, training, or demonstration approaches.

\end{abstract}

\begin{IEEEkeywords}
in-context learning, automatic speech recognition, speech LLM, attention encoder-decoder
\end{IEEEkeywords}

\section{Introduction}

In-context learning (ICL) is the ability to learn new tasks
not explicitly seen during training, typically through
a few demonstrations (i.e., example-label pairs) provided at inference time.
ICL emerged as a special capability in large language
models~(LLMs)~\cite{dong2024survey, min2022rethinking},
and has since been studied in various speech tasks, such as
speech emotion recognition~\cite{chang2024exploring, wong2025speech, ihori2025few}, speech translation~\cite{chen2024salm, tsiamas-etal-2024-speech, zhang2025mimo}, and
text-to-speech systems~\cite{nguyen2025spirit, pouw2026context}, and audio or spoken language understanding~\cite{kong2024audio,
piao2026alice, agrawal2025spoken, zhang2025mimo}.
However, studies of ICL have mostly focused on learning new {\em tasks}
and overlook the potential of using ICL to adapt to different {\em domains}.
If models are capable of learning new tasks at inference time,
they should also be able to adapt themselves to unseen domains
for the tasks seen during training.

In this work, we study ICL for adaptation, or \textbf{in-context adaptation}
for short.
A model that is able to use the demonstrations
provided at inference time to improve
performance is said to be capable of in-context adaptation.
A model capable of in-context adaptation can, for example,
adapt itself to a new accent given the examples of that accent
in the demonstrations~(accent adaptation)
or adapt its output preference given the words
in the demonstrations~(contextual biasing).
We focus on automatic speech recognition (ASR), as it is a fertile ground for adapting models to new speakers, accents, dialects, and domains. In this work, we restrict our empirical study to English ASR, enabling controlled comparison across models and datasets.
Several prior studies have approached ASR adaptation
with ICL~\cite{wang2024can, zhou2024m2r, 10890741, hsu2024smile,
omnilingual2025omnilingual, roll2025context, zheng2025ticl,
li2026multimodal}.
However, prior studies are often limited to a specific model,
such as Whisper~\cite{wang2024can, zhou2024m2r, hsu2024smile} and
Phi-4-multimodal-instruct~\cite{roll2025context, li2026multimodal, zheng2025ticl+}, or suggest that ICL
capability is elicited or reinforced through training~\cite{hsu2024smile,
pan2023cosmic, omnilingual2025omnilingual}.
Other studies focus solely on finding demonstrations
that improve performance~\cite{zhou2024m2r,
wang2024bayesian, zheng2025ticl, zheng2025ticl+}
but fail to address what exact properties in demonstrations are the most useful for adaptation.

In this work, we study whether modern ASR models can
perform in-context adaptation and what properties of demonstrations
impact the adaptation the most. 
Contrary to common belief, demonstrations in ICL for ASR do not necessarily
need to be interleaved speech and text.
We study two demonstration approaches,
\textbf{interleaved demonstration} (where a transcript is provided immediately after each of several speech examples) and \textbf{collated demonstration} (where all speech examples are provided, followed by all transcripts). 
We evaluate a diverse set of encoder-decoder models with different architectures and training recipes,
including Whisper-large-v2 (Whisper)~\cite{radford2023robust}, Canary-1B (Canary)~\cite{puvvada2024less}, Canary-Qwen-2.5B (Canary-Qwen)~\cite{nvidia_canary_qwen_2025}, Qwen3-ASR-1.7B (Qwen3-ASR)~\cite{shi2026qwen3}, Phi-4-multimodal-instruct (Phi-4-MM)~\cite{abouelenin2025phi}, and Qwen2.5-Omni-7B (Qwen2.5-Omni)~\cite{Qwen2.5-Omni}.
We find that all of them are able to perform
in-context adaptation \emph{out of the box}, i.e., without any additional training.
All models are able to adapt with collated demonstration,
whereas adapting with interleaved demonstration
is only possible for those that inherit ICL
from large language models (LLMs).

We expect the adaptation performance to depend heavily
on the demonstrations.
To understand what contributes to the adaptation,
we study in-context adaptation on L2-Arctic, 
conducting oracle experiments and controlling the speaker,
lexical content, and phonetic content in the demonstrations.
We find that sharing both lexical similarity and speaker identity leads to the most performance gains,
ranging from 25\% to 30\% WER reduction across models with collated demonstration.
To evaluate in-context adaptation
across domains, we extend our analysis to read speech that is less clean
(LibriSpeech \emph{test-other}) and
spontaneous speech (AMI).
Demonstrations sharing similar lexical content
still consistently lead to performance gains,
though the limitation of in-context adaptation becomes more obvious.

Finally, to confirm our findings, we relax the oracle experiments
by using first-pass hypotheses.
The sizable improvements are still present when using first-pass hypotheses, closing up to 60\% of the gap against the oracle.

\section{Collated and Interleaved Demonstration}
\label{sec:method}

\begin{figure}
  \centering 
  \includegraphics[width=0.9\columnwidth]{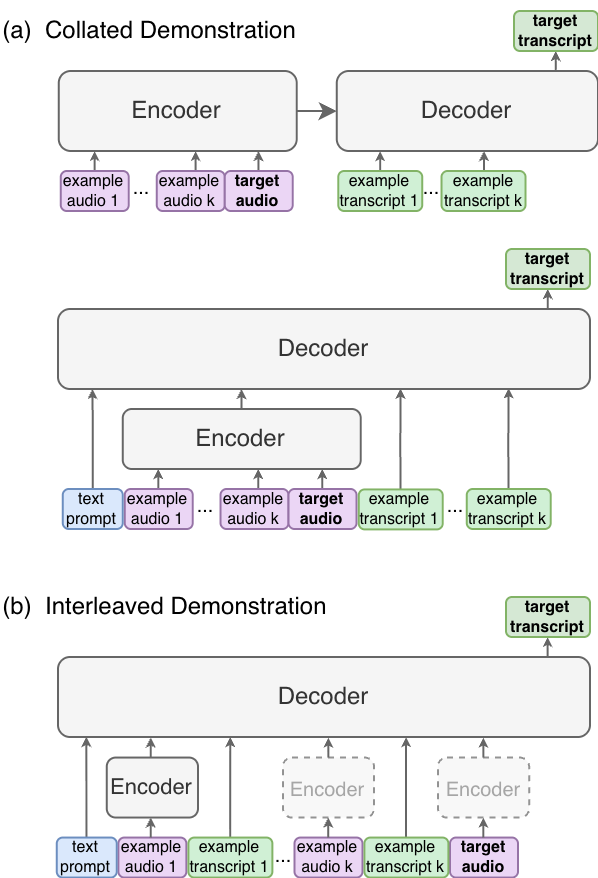} 
  \caption{Two demonstration approaches for in-context adaptation of encoder-decoder models:
    (a) \emph{Collated demonstration} for AED models (top) and for
    decoder-only/LLM-based models (bottom) and (b) \emph{Interleaved
    demonstration} for LLM-based models.}
  \label{fig:methods_illustration}
\end{figure}

When studying in-context adaptation in ASR,
we focus on encoder-decoder models,
including models with cross attention (sometimes known as attention-based encoder-decoder models, or AED~\cite{chorowski2015attention, chan2015listen, radford2023robust}),
and those without (sometimes known as decoder-only models~\cite{wu2023decoder, tsunoo2023decoder}). 
Recent LLM-based ASR models are of the second kind~\cite{omnilingual2025omnilingual, shi2026qwen3} and still require speech encoders.

As the name suggests, an encoder-decoder
model has an encoder (Enc) and decoder (Dec), and ASR is done by the generic equation 
\begin{align}
  y = \text{Dec}(\text{Enc}(x))
\end{align}
where $y$ is the output transcript when the model receives
the audio $x$ as input.

The input to the decoder is usually more
than just the encoded speech $\text{Enc}(x)$,
and the additional input is the context that
the decoder might adapt to when predicting $y$.
Formally, when using $x$ to predict $y$, the decoder
has access to additional $k$ example recordings $x_1, x_2, \dots, x_k$ 
and their respective example transcripts $y_1, y_2, \dots, y_k$.
We refer to $x$ and $y$ as the target speech and transcript,
and the pairs $(x_1, y_1), \dots, (x_k, y_k)$ as \textbf{demonstrations}.
It is common, especially in text applications, to
interleave example-label pairs as demonstrations,
but, as we show next, there are potentially better
and more natural alternatives.

\subsection{Collated demonstration}

Since an encoder-decoder model transcribes speech
in sequence, a simple approach to providing demonstrations
is, in fact, to present all the example recordings $x_1, \dots, x_k, x$ first followed
by all the example transcripts $y_1, \dots, y_k$ and let the model complete
the prediction $y$.
Formally, given the $k$ demonstration pairs
$(x_1, y_1), \dots, (x_k, y_k)$,
decoding with \textbf{collated demonstration} for AED models is defined as
\begin{align}
  y = \text{Dec}(\text{Enc}([x_1; \ldots; x_k; x]), [y_1; \ldots; y_k]),
\end{align}
where brackets ([]) and semicolons (;) are used for concatenating
vectors.
Fig.~\ref{fig:methods_illustration}(a) (top) shows
how collated demonstration is done for AED models.
The hope is that the additional transcripts $y_1, \dots, y_k$
condition the decoder (to attend to $x_1, \dots, x_k$ using attention)
when producing $y$.

Collated demonstration can be done similarly
for decoder-only architectures (such as the LLM-based ones)
using the equation
\begin{align}
  y = \text{Dec}\Big(\Big[\text{Enc}([x_1; \ldots; x_k; x]);
    y_1; \ldots; y_k \Big]\Big).
\end{align}
The only difference lies in the final concatenation.
Additional text prompts are sometimes necessary for LLM-based models.
Fig.~\ref{fig:methods_illustration}(a) (bottom) shows
how collated demonstration is done for decoder-only architectures.

In this approach, the decoder sees one long recording
that is partially transcribed, and is asked to complete the transcription.
In other words, this approach is a form of continued
decoding or prefix decoding, and is applicable
to \emph{any} encoder-decoder models.
Collated demonstration has been explored mostly with
Whisper~\cite{wang2024can, zhou2024m2r, hsu2024smile,
wang2024bayesian}.
Pan \textit{et al.}~\cite{pan2023cosmic} is the closest
to applying collated demonstration to an LLM-based ASR model,
albeit with additional instruction fine-tuning. Some work in text-to-speech apply a similar collated demonstration form for the synthesis to follow the speaker style, prosody, accent, etc.,~\cite{meng2025autoregressive, ICLR2024_ff997469, du2024cosyvoice}, where they often frame it as in-context learning or conditioning.

\subsection{Interleaved demonstration}

Many recent encoder-decoder ASR models are based on LLMs,
and part of the reason to build ASR models with LLMs is
to inherit their ICL ability.
Text-based ICL typically interleaves example-label
pairs when constructing demonstrations, and the hope
is that LLM-based ASR models can adapt themselves
when receiving speech-transcript pairs as context.
Formally, given $k$ demonstrations $(x_1, y_1), \dots, (x_k, y_k)$,
decoding with \textbf{interleaved demonstration} is defined as
\begin{align}
  y = \text{Dec}([\text{Enc}(x_1); y_1; \ldots; \text{Enc}(x_k); y_k;
    \text{Enc}(x)]).
\end{align}
Additional text prompts are almost always needed to instruct
the decoder if the model is not trained explicitly on the exact interleaved input of the same task.
Fig.~\ref{fig:methods_illustration}(b) shows
how interleaved demonstration is done for LLM-based ASR models.
Models other than those based on LLMs are not intended to be
used this way,
unless the models are explicitly trained to do so.

Interleaved demonstration has been studied in LLM-based
ASR models~\cite{roll2025context, zheng2025ticl,
li2026multimodal, omnilingual2025omnilingual}.
While effective in certain models~\cite{abouelenin2025phi,
omnilingual2025omnilingual}, interleaved demonstration
heavily depends on how models are trained
and how the text prompts are constructed.

\section{Experimental Setup}

We expect the adaptation performance to heavily depend on
the relevance between the demonstrations
and the speech to be transcribed, and we need settings
suitable for controlling phonetic, lexical, and speaker identity.
We sample six models and construct controlled demonstration settings on
three data sets, with the goal of answering
1) whether the selected models can perform in-context
adaptation and 2) what properties of the demonstrations
impact adaptation the most.

Results are based on word error rates (WER).
We follow the Open ASR Leaderboard~\cite{srivastav2025openasrleaderboardreproducible}
and use the Whisper normalizer~\cite{radford2023robust}
to normalize hypotheses and references to ensure consistent
evaluation.
Preliminary experiments show that collated demonstration can be quite sensitive to the text format or the selected examples, so to stabilize the generation, we suppress the EOS token for the first decoding step for all models to ensure the model produces at least one output token for the target audio. 

\subsection{Models and Data}

We sample six open-sourced models, mainly from the Open ASR Leaderboard,
covering attention-based encoder-decoder models (AED)
and decoder-only LLM-based models.
For LLM-based models,
some are omni models that preserve LLM capabilities and can process text,
image, and audio, while others behave strictly as an ASR model.
Table~\ref{tab:model_comparison} summarizes the selected models, their
architecture, and whether the model is designed for taking
interleaved speech and text.
Even though Canary-Qwen and Qwen3-ASR use an LLM, they are not
intended to be used for ICL.
Phi-4-MM and Qwen2.5-Omni are potentially designed with ICL in mind, since they are exposed to interleaving modality during training and preserve LLM capabilities with speech input.

\begin{table}
\centering
\caption{A summary of the selected ASR models.
  Models that are exposed to interleaved input modality during training
  are marked as intended for ICL.}
\label{tab:model_comparison}
\begin{tabular}{llc}
\toprule
\textbf{Model} & \textbf{Architecture} & \textbf{ICL} \\
\midrule
Whisper~\cite{radford2023robust} & AED & not intended \\
Canary~\cite{puvvada2024less} & AED & not intended \\
Canary-Qwen~\cite{nvidia_canary_qwen_2025} & LLM-based (ASR-only) & not intended \\
Qwen3-ASR~\cite{shi2026qwen3} & LLM-based (ASR-only) & not intended \\
Phi-4-MM~\cite{abouelenin2025phi} & LLM-based (Omni) & intended \\
Qwen2.5-Omni~\cite{Qwen2.5-Omni} & LLM-based (Omni) & intended \\ 
\bottomrule
\end{tabular}
\end{table}

To control the properties of demonstrations, we evaluate in-context
adaptation on L2-Arctic~\cite{Zhao2018L2ARCTICAN},
LibriSpeech~\cite{librispeech},
and AMI~\cite{carletta2005ami, renals2007recognition}.
L2-Arctic is a data set consisting of read
speech from non-native English speakers.
L2-Arctic is ideal for our purpose because the sentences
are phonetically balanced and the same set of sentences are read
by multiple speakers. L2-Arctic has been recently used to study in-context adaptation to speaker, accent, and lexical content~\cite{roll2025context, zheng2025ticl}.
LibriSpeech is commonly used in ASR, but more importantly,
it is useful for studying contextual biasing~\cite{tang2024improving, gong2025br, fu2023robust} 
and in our case, for controlling demonstrations of lexical similarities.
AMI consists of spontaneous speech in meetings and is
used to test the generalization beyond read speech.

\subsection{Demonstration settings}
\label{sec:settings}

To study what properties of demonstrations impact the adaptation
the most, we consider eight demonstration settings, exploring
phonetic, lexical, and speaker similarity.
We consider two extreme settings, one with demonstrations randomly sampled 
from a data set (labeled as \textbf{Random})
and one with the demonstration containing the same ground truth transcript as the target (labeled as \textbf{Gold text}).
We consider two other settings based on phonetic and lexical similarity
(labeled as \textbf{Phonetic} and \textbf{Lexical} respectively).
Given an utterance to be decoded, other utterances in the data set
are first ranked based on either phonetic similarity or lexical similarity
to the current utterance. 
The first few that are ranked top are used as demonstrations.
More formally, given the target speech $x$ and its transcript $y$,
the $k$ demonstrations are collected from a data set $S$ using
\begin{align}
(x_1, y_1), \dots, (x_k, y_k) = \mathop{\text{top-$k$}}_{(x', y') \in S}
  \text{sim}((x, y), (x', y')),
\end{align}
where $\text{sim}(\cdot, \cdot)$ is a similarity function.
Note that $y$ does not need to be the ground truth transcript
of the target speech $x$ and can be the first-pass transcription
of an ASR system, whereas the utterances in the data set $S$ always have the ground truth transcript.
For simplicity, phonetic similarity is computed based on term frequency-inverse
document frequency (TF-IDF) of phone trigrams,
and lexical similarity is based on TF-IDF of words
excluding stopwords.
Designing retrieval methods to select useful demonstrations is an active
research area~\cite{zheng2025ticl, zhou2024m2r, wang2024bayesian},
though our choice works sufficiently well as we will see in the experiments.
The demonstrations presented to a model are ordered by similarity,
with the most similar
ordered last, i.e., closest to the utterance to be decoded.
All settings are then doubled by controlling whether the demonstrations come from the same speaker or not.
Except in the \textbf{Gold text} setting, we ensure that the same text as the target is never included in the demonstrations.
For the \textbf{Gold text} setting with the same speaker, the demonstration is the target speech and its ground truth transcript.

Even though the \textbf{Phonetic} and \textbf{Lexical} settings
require access to the ground truth transcript of the target utterance,
we will relax this assumption and experiment in the same setting with first-pass transcripts to select the demonstrations.
We emphasize the importance of these oracle experiments
as they answer whether a model is capable of in-context adaptation
and to what extent the properties of demonstrations impact the adaptation.
The oracle results are also largely missing in prior work.

\section{Results}

To see whether the selected models can perform in-context adaptation,
we first test the corner cases, providing random demonstrations
or gold demonstrations.
We will then present a comprehensive comparison including
the phonetic and lexical settings,
and study generalization of in-context adaptation to other data sets.

\subsection{In-context adaptation in corner cases}

\begin{table}
\centering
\caption{WERs of in-context adaptation in corner cases on L2-Arctic.
  The symbol $\emptyset$ denotes the
  baseline without demonstrations.
  The \textbf{Random} setting includes a single random
  demonstration from a \textit{different} speaker, and the \textbf{Gold text} setting
  includes the target speech and its ground truth transcript as
  the demonstration.
}
\label{tab:conditioning_preliminary}
\begin{tabular}{lccc}
\toprule
                & $\emptyset$ & \textbf{Random} & \textbf{Gold text} \\
\midrule
Whisper         & 7.8         & 7.3    & 0.6  \\
Canary          & 6.8         & 6.7    & 2.2  \\
Canary-Qwen     & 6.7         & 6.6    & 0.2  \\
Qwen3-ASR       & 6.0         & 5.9    & 0.1  \\
Phi-4-MM        & 6.7         & 6.7    & 0.3  \\
Qwen2.5-Omni    & 5.5         & 5.6    & 0.1  \\
\bottomrule
\end{tabular}
\end{table}

\begin{figure}
  \centering
  \includegraphics[width=0.75\columnwidth]{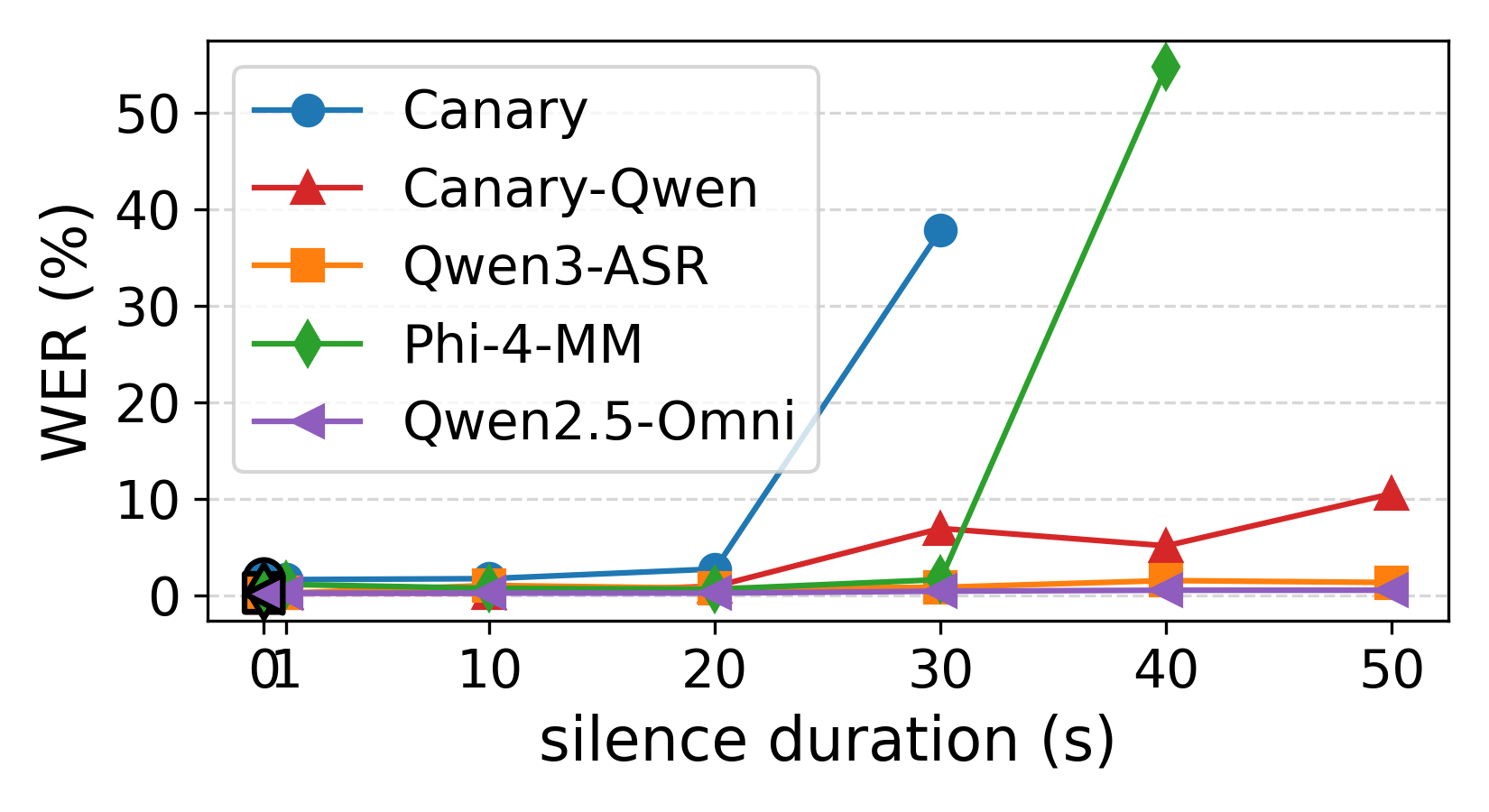}
  \caption{WERs when separating the demonstration and the
    target speech with silence in the \textbf{Gold text} setting on L2-Arctic.
    The results show the potential limit of a model's ability to handle
    long-form audio.}
  \label{fig:insert-silence}
\end{figure}

There are two corner cases tested here:
1) whether models can ignore irrelevant demonstrations
and 2) whether models can copy the ground truth
when they are given.
For the first case, one irrelevant demonstration is selected randomly from other speakers, and for the second, we provide
the identical recording and its ground truth as demonstration.
These two corner cases serve as sanity checks
that the models need to pass if we want to claim
that they can perform in-context adaptation.
We limit ourselves to collated demonstration here because,
as we have argued, it is trivially applicable to all models.

Table~\ref{tab:conditioning_preliminary} shows the results
of the two corner cases, and we see that all models
pass the test. The results with random demonstrations are
close to the results without demonstrations. All models achieve
near-perfect WERs when providing the ground truth.
Together, all models are capable of in-context adaptation
in these two corner cases.

In addition, we insert an arbitrarily long silence between the demonstration
and the speech to be decoded. A model that is capable of in-context
adaptation should be able to ignore the silence.
However, it is well known that encoder-decoder models suffer
when the input audio is too long,
beyond the typical length seen during training~\cite{chorowski2015attention, chiu2019comparison, zhang2023google}.
Fig.~\ref{fig:insert-silence} shows how models cope with the inserted silence,
and we indeed see that Canary and Phi-4-MM
(and perhaps Canary-Qwen to a certain degree) can only perform
in-context adaptation within a certain context length.
For the rest of the experiments, we respect this limit
and fit as many demonstrations as possible without breaking
the models.\footnote{In the case of L2-Arctic,
30 seconds are sufficient to fit 4 demonstrations.}

\subsection{Impact of different demonstration settings}
\label{sec:conditiong_signals}

\begin{figure*}
    \centering
    \includegraphics[width=\textwidth]{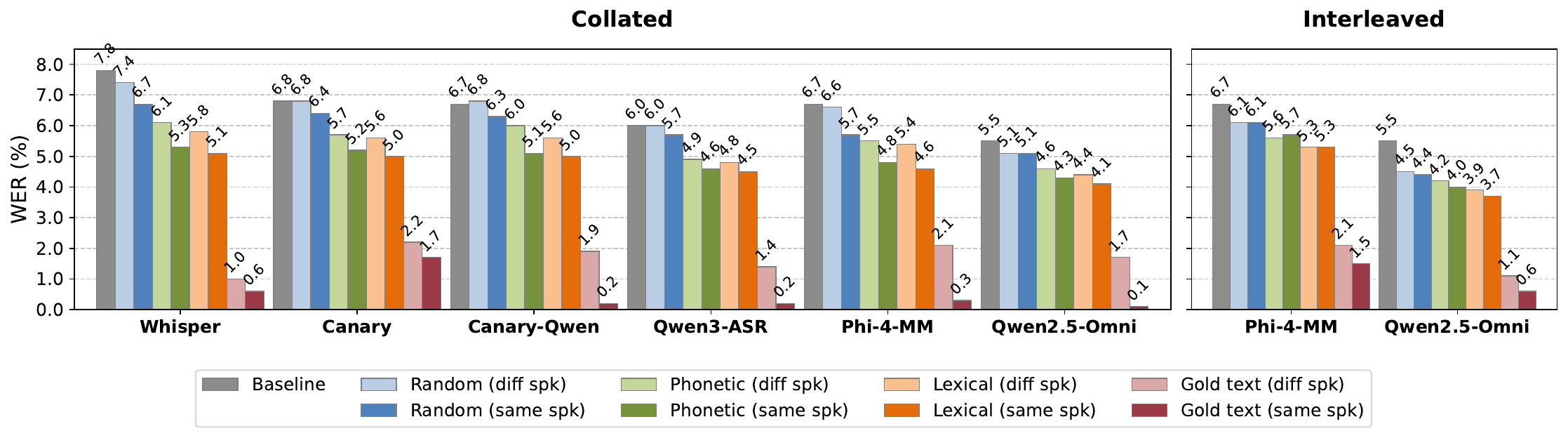}
    \caption{WERs of in-context adaptation on L2-Arctic for different demonstration settings. We use 4 demonstrations for all settings except the \textbf{Gold text}, which only has 1 demonstration.
    }
    \label{fig:hierarchy_l2arctic}
\end{figure*}

\begin{figure*}
    \centering
    \includegraphics[width=\textwidth]{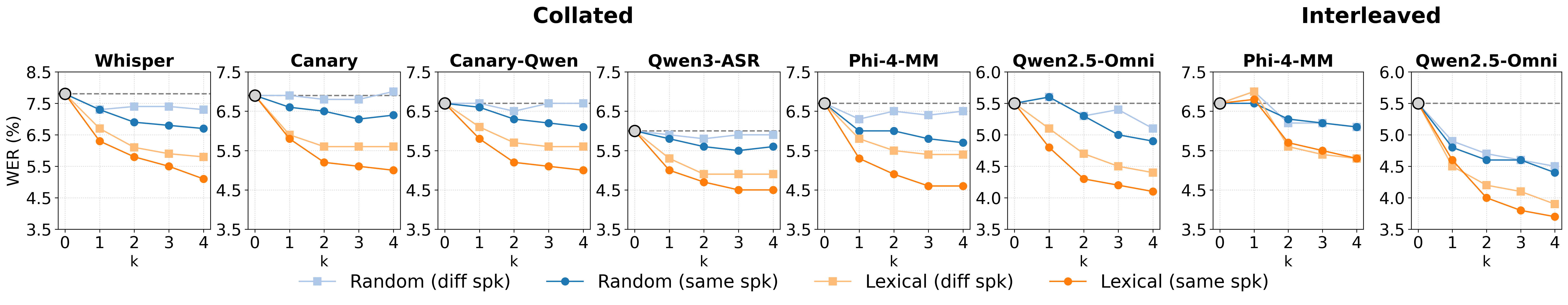}
    \caption{The \emph{absolute} WER on L2-Arctic under the \textbf{Random} and \textbf{Lexical} settings as we increase the number of demonstrations from 1 to 4.
    }
    \label{fig:scaling_k_l2arctic}
\end{figure*}

To quantify how different sources of information in speech
contribute to adaptation, we compare demonstrations selected
based on phonetic and lexical similarities, in addition to the
two corner cases.
We also extend the number of demonstrations up to four.
After controlling the speakers, we have a total of eight demonstration
settings.

Fig.~\ref{fig:hierarchy_l2arctic} shows how much models are able to adapt
based on four demonstrations of different types.
First, except \textbf{Random} with different speakers,
all demonstration settings lead to improved WERs. 
In particular, all models have no trouble getting near-perfect WERs
with the \textbf{Gold text} demonstrations.
While \textbf{Gold text} is not realistic,
it does show the importance of speakers in the demonstrations.
The \textbf{Gold text} WERs are slightly higher when the speakers are different from the target speech.

We then look at demonstration settings that do not contain duplicate text to the target transcript. Selecting random utterances from the same speaker achieves a
small but consistent gain.
Demonstrations selected based on phonetic and lexical similarity (especially
from the same speaker) lead to the most improvement, with lexical similarity slightly better. 
This trend holds across all models. 

Notably, the smaller AED models, Whisper and Canary-1B,
achieve similar relative WER
reduction to larger models equipped with pretrained LLM decoders.
This suggests that the adaptation capability does not inherently depend on
the use of a pre-trained LLM.

The amount of adaptation for different numbers of demonstrations
is shown in Fig.~\ref{fig:scaling_k_l2arctic}.
We only contrast the \textbf{Random} and the \textbf{Lexical} setting
for clarity.
Overall, from one to four examples, the more demonstrations we provide, the lower the WER.
The first two demonstrations bring the most improvement, and
adding more demonstrations diminishes the return.

\subsection{Comparing collated and interleaved demonstration}
\label{sec:prompt_structure_compare}

\begin{figure*}
    \centering
    \includegraphics[width=\textwidth]{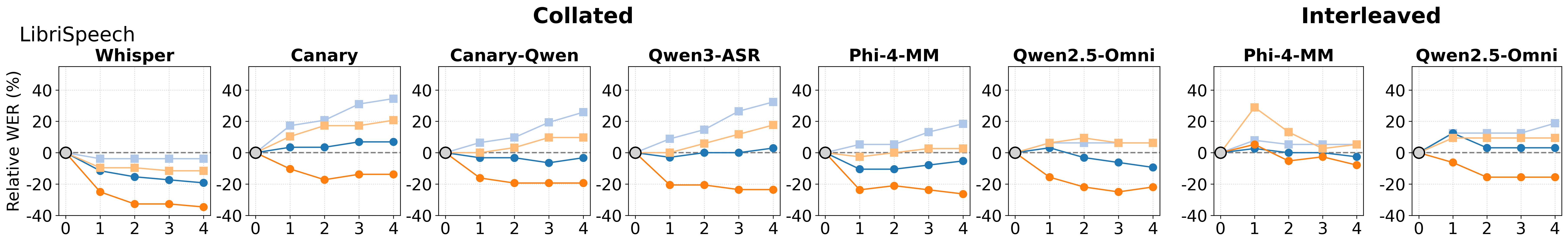} \\
    \includegraphics[width=\textwidth]{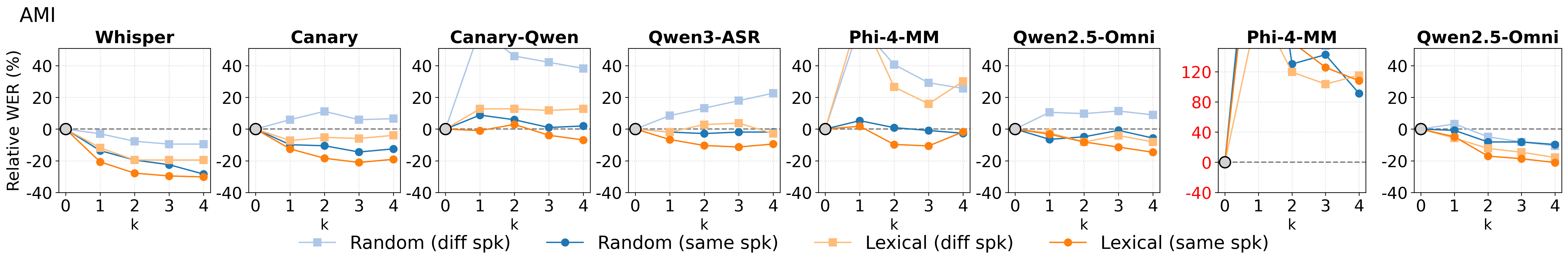} 
    \caption{\textit{Relative} WER reduction of in-context adaptation against no adaptation on LibriSpeech \textit{test-other} (top) and AMI test split from Open ASR Leaderboard (bottom). We use relative reduction because the WERs span a larger dynamic range across models compared to L2-Arctic.
    }
    \label{fig:ls_scaling}
\end{figure*}

Comparing Phi-4-MM and Qwen2.5-Omni in Fig.~\ref{fig:hierarchy_l2arctic} and Fig.~\ref{fig:ls_scaling},
we see that both models are able to adapt based on
collated and interleaved demonstration.
For interleaved demonstration,
we follow the exact same instruction prompt in Roll \textit{et al.}~\cite{roll2025context}
for both Phi-4-MM and Qwen2.5-Omni.
Note that prompt engineering is required for interleaved demonstration for both models. Without this instruction prompt, we observe much weaker adaptation, especially if only a single example is provided.

Even though Canary-Qwen and Qwen3-ASR are based on LLMs,
they are not intended to take interleaved input and
we also do not observe any adaptation effect when providing interleaved demonstrations for these two models.
A key advantage of collated demonstration is that it is
applicable to all models, requires no modification
to the text instruction and task prompt, and works
out of the box.

\subsection{In-context adaptation on different domains}
\label{sec:conditioning_diff_data_domains}

We extend our analysis to LibriSpeech (\textit{test-other}) and AMI.
We again contrast \textbf{Random} and \textbf{Lexical} for clarity.
The top row of Fig.~\ref{fig:ls_scaling} shows the results
on LibriSpeech \textit{test-other}.
We see that the overall improvement is smaller on these two
data sets compared to L2-Arctic.
In particular, providing utterances just from the same speaker
is overall less effective.
Using lexically similar utterances from the same speaker
continues to be the best strategy, achieving
at least 20\% relative WER reduction with collated demonstration.

Results on AMI are shown in the bottom row of Fig.~\ref{fig:ls_scaling}.
Since AMI is the more challenging data set consisting of
spontaneous speech, the adaptation is again less effective.
In fact, Phi-4-MM completely fails under interleaved demonstration, where we found the WER collapses due to severe copying from the demonstration transcripts,
though Qwen2.5-Omni adapts just fine.
The conclusion is largely the same as in LibriSpeech,
lexically similar demonstrations from the same speaker
achieve the most improvement.

Interestingly, while LLM-based models show marginal gains under
the \textbf{Lexical} setting, Whisper shows the
largest gains, almost closing the WER gap to the LLM-based ASR models. This
again suggests that the effectiveness of in-context adaptation does not necessarily require the use of an LLM.

\section{Second-pass in-context adaptation}
\label{sec:proxy}

Our oracle results, though promising, assume
access to the ground truth transcript of the target speech.
To evaluate the gains in practical scenarios,
we relax the assumption and use first-pass
transcriptions to find demonstrations of
the same speaker and of high lexical similarity.
For simplicity, we use the same TF-IDF in the \textbf{Lexical}
setting in \S\ref{sec:settings}.
We decide to measure similarity using ASR transcripts
but still use gold transcriptions in demonstrations.
Formally, assuming that we have a
data set $S$ with gold transcriptions,
we construct the demonstrations with
\begin{align}
(x_1, y_1), \dots, (x_k, y_k) = \mathop{\text{top-$k$}}_{(x', y') \in S}
  \text{TF-IDF}(T(x), T(x')),
\end{align}
where $T$ is an ASR system and $T(x)$ is the first-pass transcript
of the target speech $x$.
Note that the ground truth transcript of the example $y'$ is \emph{not} used in computing TF-IDF. 

Results with this \emph{second-pass} in-context adaptation on L2-Arctic
are shown in Table \ref{tab:proxy}.
We see a further 3\% to 12\% relative WER reduction
over the \textbf{Random} setting,
corresponding to closing 25\% to 60\% of the gaps against the oracle.
The second-pass approach is similar to those based on
retrieval~\cite{zheng2025ticl, zhou2024m2r, wang2024bayesian}.
While the retrieval approaches are complementary to ours, a simple
second pass already brings a significant improvement.

\begin{table}
\centering
\caption{WERs of second-pass in-context adaptation on L2-Arctic in the \textbf{Random} and \textbf{Lexical} settings using 4 demonstrations from the \emph{same} speakers.
The \emph{first pass} column uses the first-pass transcription for selecting demonstrations, whereas the \emph{oracle} column selects demonstrations with the ground truth transcripts.}
\resizebox{\columnwidth}{!}{%
\begin{tabular}{lcccc}
\toprule
\multirow{2}[2]{*}{Model} & \multirow{2}[2]{*}{$\emptyset$} & \multirow{2}[2]{*}{\textbf{Random}} & \multicolumn{2}{c}{\textbf{Lexical}} \\
\cmidrule{4-5}
& &  & first pass  & oracle \\
\midrule
\textbf{Collated demonstration} \\
\hspace{0.5em} Whisper       & 7.8 & 6.7 & 6.0 & 5.1 \\ 
\hspace{0.5em} Canary        & 6.8 & 6.1 & 5.5 & 5.0 \\ 
\hspace{0.5em} Canary-Qwen   & 6.7 & 6.2 & 5.5 & 5.0 \\
\hspace{0.5em} Qwen3-ASR     & 6.0 & 5.6 & 5.0 & 4.5 \\
\hspace{0.5em} Phi-4-MM      & 6.7 & 5.7 & 5.4 & 4.6 \\ 
\hspace{0.5em} Qwen2.5-Omni  & 5.5 & 5.1 & 4.5 & 4.1 \\
\midrule                                                

\textbf{Interleaved demonstration} \\                   
\hspace{0.5em} Phi-4-MM      & 6.7 & 6.1 & 5.9 & 5.3 \\ 
\hspace{0.5em} Qwen2.5-Omni  & 5.5 & 4.4 & 4.2 & 3.7 \\ 
\bottomrule
\end{tabular}
}
\label{tab:proxy}
\end{table}

\section{Discussion}

Though both collated and interleaved demonstration can achieve
in-context adaptation, their underlying mechanism might be different.
Collated demonstration operates as one ASR task,
while interleaved demonstration requires additional understanding
of text instructions and the interleaving format.
This fact makes collated demonstration widely applicable
to any encoder-decoder ASR system,
and makes interleaved demonstration sensitive
to exact wording of the text prompts.
Partly because of this, the performance of interleaved demonstration
varies a lot more across data sets compared to that of collated demonstration.

Both demonstration approaches, however, are limited
by the context length of the model.
It is well known that ASR systems suffer from long-form audio or the audio length beyond training~\cite{chorowski2015attention, chiu2019comparison, zhang2023google}.
The problem is partially addressed by the use of an LLM, as
long-context modeling~\cite{yang2025qwen3, abouelenin2025phi} are now an integrated
part of training LLMs.
The gains of using more demonstrations do diminish in our experiments,
so this limitation might not be as significant.

\section{Conclusion}

In this work, we study in-context adaptation in ASR with two approaches, collated and interleaved demonstration.
We show that, without any training, in-context adaptation achieves significant gains across a diverse set of models and settings.

The oracle experiments not only establish the fact that all models tested are capable of in-context adaptation but also show the limit of what in-context adaptation can achieve.
Notably, smaller AED models adapt as effectively as larger LLM-based models, suggesting that in-context adaptation in ASR does not inherently depend on an LLM decoder.

Between the two demonstration approaches, collated demonstration is trivially applicable to any encoder-decoder ASR system, while interleaved demonstration is only applicable to certain models, requiring specific text prompts. 

We also show that our approaches are practical, relaxing the oracle results with first-pass hypotheses while maintaining a large proportion of the improvements.
Future work includes a deeper understanding of the mechanism behind in-context adaptation, designing experiments to disentangle the contribution of text and audio, developing retrieval approaches for finding better demonstrations, and extending to other settings, such as noise robustness and low-resource ASR. 

\section{Generative AI Use Disclosure}
Generative AI tools (Claude) were only used for language and grammar polishing of the manuscript. All technical content, experimental design, and results were produced by the authors.

\bibliographystyle{IEEEtran}
\bibliography{refs}

@article{srivastav2025openasrleaderboardreproducible,
  title={Open {ASR} Leaderboard: Towards Reproducible and Transparent Multilingual and Long-Form Speech Recognition Evaluation}, 
  author={Vaibhav Srivastav and Steven Zheng and Eric Bezzam and Eustache Le Bihan and Nithin Koluguri and Piotr {\.Z}elasko and Somshubra Majumdar and Adel Moumen and Sanchit Gandhi},
  year={2025},
  journal={arXiv preprint arXiv:2510.06961}
}

@inproceedings{puvvada2024less,
  title={Less is more: Accurate speech recognition \& translation without web-scale data},
  author={Puvvada, Krishna C and {\.Z}elasko, Piotr and Huang, He and Hrinchuk, Oleksii and Koluguri, Nithin Rao and Dhawan, Kunal and Majumdar, Somshubra and Rastorgueva, Elena and Chen, Zhehuai and Lavrukhin, Vitaly and others},
  booktitle={Interspeech},
  year={2024}
}

@inproceedings{roll2025context,
  title={In-context learning boosts speech recognition via human-like adaptation to speakers and language varieties},
  author={Roll, Nathan and Graham, Calbert and Tatsumi, Yuka and Nguyen, Kim Tien and Sumner, Meghan and Jurafsky, Dan},
  booktitle={EMNLP},
  year={2025}
}

@inproceedings{zheng2025ticl,
  title={{TICL}: Text-Embedding {KNN} For Speech In-Context Learning Unlocks Speech Recognition Abilities of Large Multimodal Models},
  author={Zheng, Haolong and Yegorova, Yekaterina and Hasegawa-Johnson, Mark},
  booktitle={ICASSP},
  year={2025}
}

@article{abouelenin2025phi,
  title={Phi-4-mini technical report: Compact yet powerful multimodal language models via mixture-of-{L}o{RA}s},
  author={Abouelenin, Abdelrahman and Ashfaq, Atabak and Atkinson, Adam and Awadalla, Hany and Bach, Nguyen and Bao, Jianmin and Benhaim, Alon and Cai, Martin and Chaudhary, Vishrav and Chen, Congcong and others},
  journal={arXiv preprint arXiv:2503.01743},
  year={2025}
}

@inproceedings{wang2024can,
  title={Can {Whisper} perform speech-based in-context learning?},
  author={Wang, Siyin and Yang, Chao-Han and Wu, Ji and Zhang, Chao},
  booktitle={ICASSP},
  year={2024}
}

@inproceedings{zhou2024m2r,
  title={{M2R-Whisper}: Multi-stage and Multi-scale retrieval augmentation for enhancing {W}hisper},
  author={Zhou, Jiaming and Zhao, Shiwan and He, Jiabei and Wang, Hui and Zeng, Wenjia and Chen, Yong and Sun, Haoqin and Kong, Aobo and Qin, Yong},
  booktitle={ICASSP},
  year={2025}
}

@inproceedings{hsu2024smile,
  title={{SMILE}: Speech meta in-context learning for low-resource language automatic speech recognition},
  author={Hsu, Ming-Hao and Lee, Hung-yi},
  booktitle={ASRU},
  year={2025}
}

@inproceedings{wang2024bayesian,
  title={Bayesian example selection improves in-context learning for speech, text and visual modalities},
  author={Wang, Siyin and Yang, Chao-Han Huck and Wu, Ji and Zhang, Chao},
  booktitle={EMNLP},
  year={2024}
}

@misc{omnilingual2025omnilingual,
      title={Omnilingual {ASR}: Open-Source Multilingual Speech Recognition for 1600+ Languages},
      author={\text{Omnilingual ASR team} and Gil Keren and Artyom Kozhevnikov and Yen Meng and Christophe Ropers and Matthew Setzler and Skyler Wang and Ife Adebara and Michael Auli and Can Balioglu and Kevin Chan and Chierh Cheng and Joe Chuang and Caley Droof and Mark Duppenthaler and Paul-Ambroise Duquenne and Alexander Erben and Cynthia Gao and Gabriel Mejia Gonzalez and Kehan Lyu and Sagar Miglani and Vineel Pratap and Kaushik Ram Sadagopan and Safiyyah Saleem and Arina Turkatenko and Albert Ventayol-Boada and Zheng-Xin Yong and Yu-An Chung and Jean Maillard and Rashel Moritz and Alexandre Mourachko and Mary Williamson and Shireen Yates},
      year={2025},
      url={https://arxiv.org/abs/2511.09690},
}

@inproceedings{chan2015listen,
  title={Listen, attend and spell: A neural network for large vocabulary conversational speech recognition},
  author={Chan, William and Jaitly, Navdeep and Le, Quoc V and Vinyals, Oriol},
  booktitle={ICASSP},
  year={2016}
}

@inproceedings{radford2023robust,
  title={Robust speech recognition via large-scale weak supervision},
  author={Radford, Alec and Kim, Jong Wook and Xu, Tao and Brockman, Greg and McLeavey, Christine and Sutskever, Ilya},
  booktitle={ICML},
  year={2023},
}

@inproceedings{Zhao2018L2ARCTICAN,
  title={{L2-ARCTIC}: A Non-native English Speech Corpus},
  author={Guanlong Zhao and Sinem Sonsaat and Alif Silpachai and Ivana Lucic and Evgeny Chukharev-Hudilainen and John M. Levis and Ricardo Gutierrez-Osuna},
  booktitle={Interspeech},
  year={2018},
}

@article{shi2026qwen3,
  title={Qwen3-{ASR} Technical Report},
  author={Shi, Xian and Wang, Xiong and Guo, Zhifang and Wang, Yongqi and Zhang, Pei and Zhang, Xinyu and Guo, Zishan and Hao, Hongkun and Xi, Yu and Yang, Baosong and others},
  journal={arXiv preprint arXiv:2601.21337},
  year={2026}
}

@article{pouw2026context,
  title={In-Context Learning in Speech Language Models: Analyzing the Role of Acoustic Features, Linguistic Structure, and Induction Heads},
  author={Pouw, Charlotte and Mohebbi, Hosein and Alishahi, Afra and Zuidema, Willem},
  journal={arXiv preprint arXiv:2604.06356},
  year={2026}
}

@article{yang2025qwen3,
  title={Qwen3 technical report},
  author={Yang, An and Li, Anfeng and Yang, Baosong and Zhang, Beichen and Hui, Binyuan and Zheng, Bo and Yu, Bowen and Gao, Chang and Huang, Chengen and Lv, Chenxu and others},
  journal={arXiv preprint arXiv:2505.09388},
  year={2025}
}

@inproceedings{pan2023cosmic,
  title={{COSMIC}: Data efficient instruction-tuning for speech in-context learning},
  author={Pan, Jing and Wu, Jian and Gaur, Yashesh and Sivasankaran, Sunit and Chen, Zhuo and Liu, Shujie and Li, Jinyu},
  year={2024},
  booktitle={Interspeech}
}

@inproceedings{chen2024salm,
  title={{SALM}: Speech-augmented language model with in-context learning for speech recognition and translation},
  author={Chen, Zhehuai and Huang, He and Andrusenko, Andrei and Hrinchuk, Oleksii and Puvvada, Krishna C and Li, Jason and Ghosh, Subhankar and Balam, Jagadeesh and Ginsburg, Boris},
  booktitle={ICASSP},
  year={2024}
}

@inproceedings{li2026multimodal,
  title={Multimodal In-context Learning for {ASR} of Low-resource Languages},
  author={Li, Zhaolin and Niehues, Jan},
  booktitle={Findings of the ACL},
  year={2026}
}

@inproceedings{ihori2025few,
  title={Few-shot Personalization via In-Context Learning for Speech Emotion Recognition based on Speech-Language Model},
  author={Ihori, Mana and Yamane, Taiga and Kawata, Naotaka and Makishima, Naoki and Tanaka, Tomohiro and Suzuki, Satoshi and Orihashi, Shota and Masumura, Ryo},
  booktitle={ASRU},
  year={2025}
}

@article{piao2026alice,
  title={{ALICE}: A Multifaceted Evaluation Framework of Large Audio-Language Models' In-Context Learning Ability},
  author={Piao, Yen-Ting and Liao, Jay Chiehen and Chien, Wei-Tang and Ogimoto, Toshiki and Chen, Shang-Tse and Chen, Yun-Nung and Lee, Chun-Yi and Lo, Shao-Yuan},
  journal={arXiv preprint arXiv:2603.20433},
  year={2026}
}

@inproceedings{wong2025speech,
  title={Speech in-context learning of paralinguistic tasks},
  author={Wong, Jeremy HM and Huzaifah, Muhammad and Chen, Nancy F and Aw, Ai Ti},
  booktitle={ASRU},
  year={2025},
}

@inproceedings{chang2024exploring,
  title={Exploring In-Context Learning of Textless Speech Language Model for Speech Classification Tasks},
  author={Chang, Kai-Wei and Hsu, Ming-Hao and Li, Shang-Wen and Lee, Hung-yi},
  booktitle={Interspeech},
  year={2024}
}

@inproceedings{gong2025br,
  title={{BR-ASR}: Efficient and scalable bias retrieval framework for contextual biasing {ASR} in speech {LLM}},
  author={Gong, Xun and Lv, Anqi and Wang, Zhiming and Zhu, Huijia and Qian, Yanmin},
  booktitle={Interspeech},
  year={2025}
}

@inproceedings{tang2024improving,
  title={Improving {ASR} contextual biasing with guided attention},
  author={Tang, Jiyang and Kim, Kwangyoun and Shon, Suwon and Wu, Felix and Sridhar, Prashant},
  booktitle={ICASSP},
  year={2024}
}

@inproceedings{dong2024survey,
  title={A survey on in-context learning},
  author={Dong, Qingxiu and Li, Lei and Dai, Damai and Zheng, Ce and Ma, Jingyuan and Li, Rui and Xia, Heming and Xu, Jingjing and Wu, Zhiyong and Chang, Baobao and others},
  booktitle={EMNLP},
  year={2024}
}

@inproceedings{min2022rethinking,
  title={Rethinking the role of demonstrations: What makes in-context learning work?},
  author={Min, Sewon and Lyu, Xinxi and Holtzman, Ari and Artetxe, Mikel and Lewis, Mike and Hajishirzi, Hannaneh and Zettlemoyer, Luke},
  booktitle={EMNLP},
  year={2022}
}

@article{Qwen2.5-Omni,
  title={Qwen2.5-{O}mni Technical Report},
  author={Jin Xu and Zhifang Guo and Jinzheng He and Hangrui Hu and Ting He and Shuai Bai and Keqin Chen and Jialin Wang and Yang Fan and Kai Dang and Bin Zhang and Xiong Wang and Yunfei Chu and Junyang Lin},
  journal={arXiv preprint arXiv:2503.20215},
  year={2025}
}

@inproceedings{librispeech,
  author={Panayotov, Vassil and Chen, Guoguo and Povey, Daniel and Khudanpur, Sanjeev},
  booktitle={ICASSP}, 
  title={{LibriSpeech}: An {ASR} corpus based on public domain audio books}, 
  year={2015}
}

@inproceedings{renals2007recognition,
  title={Recognition and understanding of meetings the {AMI} and {AMIDA} projects},
  author={Renals, Steve and Hain, Thomas and Bourlard, Herv{\'e}},
  booktitle={ASRU},
  year={2007}
}

@inproceedings{carletta2005ami,
  title={The {AMI} meeting corpus: A pre-announcement},
  author={Carletta, Jean and Ashby, Simone and Bourban, Sebastien and Flynn, Mike and Guillemot, Mael and Hain, Thomas and Kadlec, Jaroslav and Karaiskos, Vasilis and Kraaij, Wessel and Kronenthal, Melissa and others},
  booktitle={International Workshop on Machine Learning for Multimodal Interaction},
  year={2005}
}

@inproceedings{meng2025autoregressive,
  title={Autoregressive speech synthesis without vector quantization},
  author={Meng, Lingwei and Zhou, Long and Liu, Shujie and Chen, Sanyuan and Han, Bing and Hu, Shujie and Liu, Yanqing and Li, Jinyu and Zhao, Sheng and Wu, Xixin and others},
  booktitle={ACL},
  year={2025}
}

@article{tsunoo2023decoder,
  title={Decoder-only architecture for speech recognition with {CTC} prompts and text data augmentation},
  author={Tsunoo, Emiru and Futami, Hayato and Kashiwagi, Yosuke and Arora, Siddhant and Watanabe, Shinji},
  journal={arXiv preprint arXiv:2309.08876},
  year={2023}
}

@inproceedings{wu2023decoder,
  title={On decoder-only architecture for speech-to-text and large language model integration},
  author={Wu, Jian and Gaur, Yashesh and Chen, Zhuo and Zhou, Long and Zhu, Yimeng and Wang, Tianrui and Li, Jinyu and Liu, Shujie and Ren, Bo and Liu, Linquan and others},
  booktitle={ASRU},
  year={2023}
}

@misc{nvidia_canary_qwen_2025,
  title        = {{NVIDIA NeMo Canary-Qwen-2.5B}},
  author       = {{NVIDIA}},
  year         = {2025},
  howpublished = {\url{https://huggingface.co/nvidia/canary-qwen-2.5b}},
  note         = {{Hugging Face} model card. Accessed: 2026-06-08}
}

@inproceedings{fu2023robust,
  title={Robust acoustic and semantic contextual biasing in neural transducers for speech recognition},
  author={Fu, Xuandi and Sathyendra, Kanthashree Mysore and Gandhe, Ankur and Liu, Jing and Strimel, Grant P and McGowan, Ross and Mouchtaris, Athanasios},
  booktitle={ICASSP},
  year={2023}
}

@inproceedings{chorowski2015attention,
 title = {Attention-Based Models for Speech Recognition},
 author = {Chorowski, Jan K and Bahdanau, Dzmitry and Serdyuk, Dmitriy and Cho, Kyunghyun and Bengio, Yoshua},
 booktitle = {Advances in Neural Information Processing Systems},
 year = {2015}
}

@inproceedings{zheng2025ticl+,
  title={{TICL+}: A Case Study On Speech In-Context Learning for Children's Speech Recognition},
  author={Zheng, Haolong and Yegorova, Yekaterina and Hasegawa-Johnson, Mark},
  booktitle={ASRU Satellite Workshop---AI for Children's Speech and Language},
  year={2025}
}

@article{zhang2023google,
  title={Google {USM}: Scaling automatic speech recognition beyond 100 languages},
  author={Zhang, Yu and Han, Wei and Qin, James and Wang, Yongqiang and Bapna, Ankur and Chen, Zhehuai and Chen, Nanxin and Li, Bo and Axelrod, Vera and Wang, Gary and others},
  journal={arXiv preprint arXiv:2303.01037},
  year={2023}
}

@inproceedings{chiu2019comparison,
  title={A comparison of end-to-end models for long-form speech recognition},
  author={Chiu, Chung-Cheng and Han, Wei and Zhang, Yu and Pang, Ruoming and Kishchenko, Sergey and Nguyen, Patrick and Narayanan, Arun and Liao, Hank and Zhang, Shuyuan and Kannan, Anjuli and others},
  booktitle={ASRU},
  year={2019}
}

@INPROCEEDINGS{10890741,
  author={Cheng, Jian and Nguyen, Sam},
  booktitle={ICASSP}, 
  title={Speech Few-Shot Learning for Language Learners’ Speech Recognition}, 
  year={2025}
}

@inproceedings{agrawal2025spoken,
  title={Spoken language understanding on unseen tasks with in-context learning},
  author={Agrawal, Neeraj and Ganapathy, Sriram},
  booktitle={Interspeech},
  year={2025}
}

@inproceedings{kong2024audio,
  title={{Audio Flamingo}: A novel audio language model with few-shot learning and dialogue abilities},
  author={Kong, Zhifeng and Goel, Arushi and Badlani, Rohan and Ping, Wei and Valle, Rafael and Catanzaro, Bryan},
  booktitle={ICML},
  year={2024}
}

@inproceedings{ICLR2024_ff997469,
  title = {{Mega-TTS} 2: Boosting Prompting Mechanisms for Zero-Shot Speech Synthesis},
  author = {Jiang, Ziyue and Liu, Jinglin and Ren, Yi and He, Jinzheng and Ye, Zhenhui and Ji, Shengpeng and Yang, Qian and Zhang, Chen and Wei, Pengfei and Wang, Chunfeng and Yin, Xiang and MA, Zejun and Zhao, Zhou},
  booktitle = {ICLR},
  year = {2024}
}

@article{du2024cosyvoice,
  title={Cosyvoice 2: Scalable streaming speech synthesis with large language models},
  author={Du, Zhihao and Wang, Yuxuan and Chen, Qian and Shi, Xian and Lv, Xiang and Zhao, Tianyu and Gao, Zhifu and Yang, Yexin and Gao, Changfeng and Wang, Hui and others},
  journal={arXiv preprint arXiv:2412.10117},
  year={2024}
}

@inproceedings{tsiamas-etal-2024-speech,
    title = "Speech Is More than Words: Do Speech-to-Text Translation Systems Leverage Prosody?",
    author = "Tsiamas, Ioannis  and
      Sperber, Matthias  and
      Finch, Andrew  and
      Garg, Sarthak",
    booktitle = "Conference on Machine Translation",
    year = "2024",
}

@article{zhang2025mimo,
  title={MiMo-Audio: Audio Language Models are Few-Shot Learners},
  author={Zhang, Dong and Wang, Gang and Xue, Jinlong and Fang, Kai and Zhao, Liang and Ma, Rui and Ren, Shuhuai and Liu, Shuo and Guo, Tao and Zhuang, Weiji and others},
  journal={arXiv preprint arXiv:2512.23808},
  year={2025}
}

@article{nguyen2025spirit,
  title={Spirit-{LM}: Interleaved spoken and written language model},
  author={Nguyen, Tu Anh and Muller, Benjamin and Yu, Bokai and Costa-Jussa, Marta R and Elbayad, Maha and Popuri, Sravya and Ropers, Christophe and Duquenne, Paul-Ambroise and Algayres, Robin and Mavlyutov, Ruslan and others},
  journal={Transactions of ACL},
  year={2025}
}

\end{document}